\documentclass[runningheads]{llncs}

\usepackage{eccv}

\usepackage{eccvabbrv}

\usepackage{graphicx}
\usepackage{booktabs}
\usepackage{multirow}
\usepackage[table]{xcolor}
\usepackage{wrapfig}
\usepackage[percent]{overpic}

\usepackage[accsupp]{axessibility}  

\usepackage{hyperref}

\usepackage{orcidlink}

\begin{document}

\title{OrthoTryOn: Geometric Orthogonalization for Conflict-Free Unified Fashion Generation} 

\titlerunning{OrthoTryOn}

\author{Zhaotong Yang\inst{1} \and
Ying Tai\inst{2} \thanks{Corresponding authors.} \and
Jiahui Zhan\inst{3} \and
Yu Zheng\inst{1} \and \\
Jianjun Qian\inst{1} \and
Jian Yang\inst{1,2} \textsuperscript{$\star$}}

\authorrunning{Yang et al.}

\institute{PCA Lab, School of Computer Science and Engineering, Nanjing University of Science and Technology \and
PCA Lab, School of Intelligence Science and Technology, Nanjing University \and
Shanghai Jiao Tong University}
\maketitle

\vspace{-6mm}
\begin{figure}
	\centering
    \includegraphics[width=0.98\linewidth]{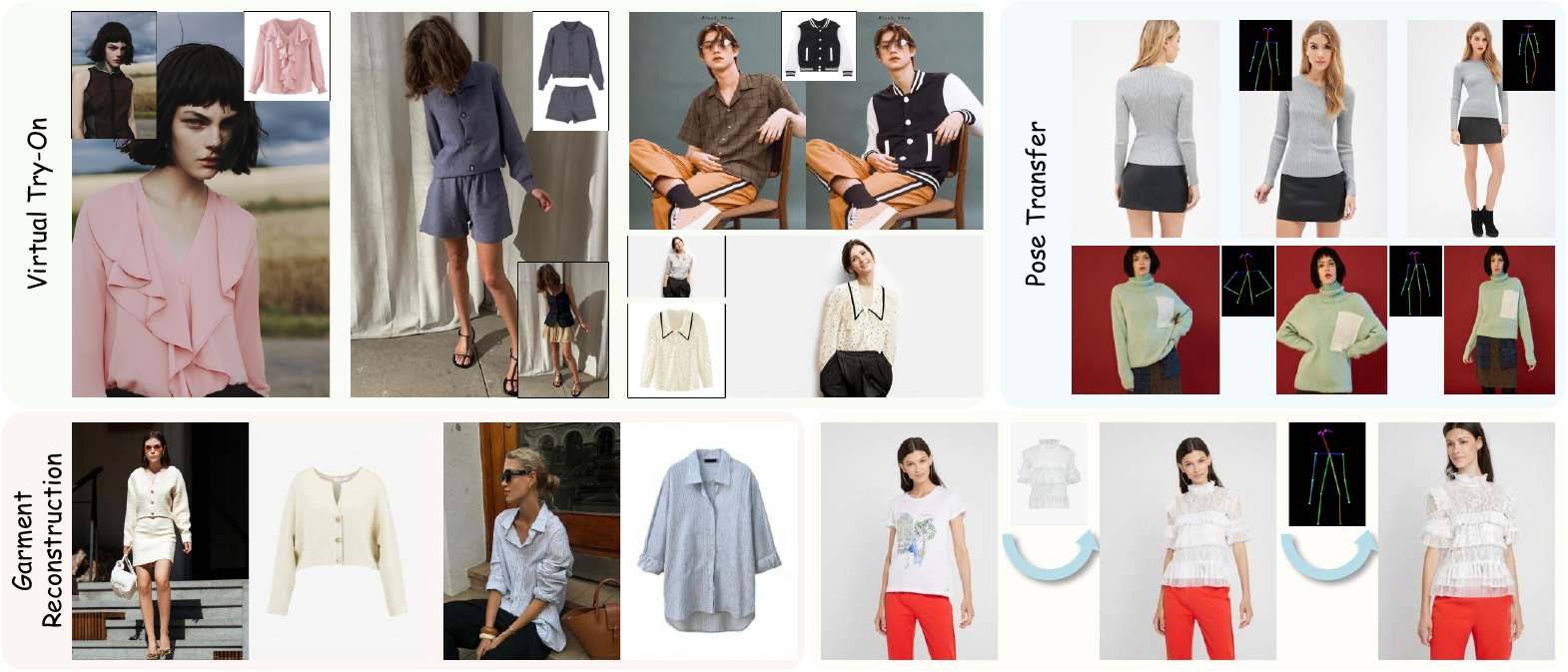}
    \caption{We present OrthoTryOn, a unified generalist model capable of handling diverse fashion tasks within a single architecture, including virtual try-on, garment reconstruction, and pose transfer. The shared architecture also naturally supports sequential editing by chaining task-specific conditions.}
    \label{fig:teaser}
\end{figure}

\vspace{-10mm}\begin{abstract}
Unified fashion generation integrates tasks like virtual try-on and garment reconstruction into a single model to reduce task-specific adaptation costs. However, naive parameter sharing across semantically distinct tasks induces negative transfer through severe inter-task gradient conflict. We propose OrthoTryOn, a unified framework mitigating this interference within a shared Low-Rank Adaptation (LoRA) module. Its Orthogonal Subspace Projection (OSP) applies task-specific orthogonal rotations to bottleneck features, mapping them into decorrelated coordinate frames. To address residual semantic coupling at inference time, we further propose Fisher-guided Negative Guidance (FNG), a parameter-free strategy that utilizes diagonal Fisher information to quantify inter-task sensitivity overlap and explicitly repels generation trajectories from the most confusable task via Classifier-Free Guidance. Extensive experiments demonstrate that OrthoTryOn avoids the severe performance degradation typical of naive unified training and even surpasses independently trained task-specific models, achieving state-of-the-art results across multiple benchmarks while generalizing robustly across diverse diffusion backbones. Code is available at \url{https://github.com/NJU-PCALab/OrthoTryOn}.

  \keywords{Virtual Try-On \and Unified Fashion Generation \and Diffusion Model}
\end{abstract}

\section{Introduction}\label{sec:intro}

\begin{figure}[t]
	\centering
    \begin{overpic}[width=\linewidth]{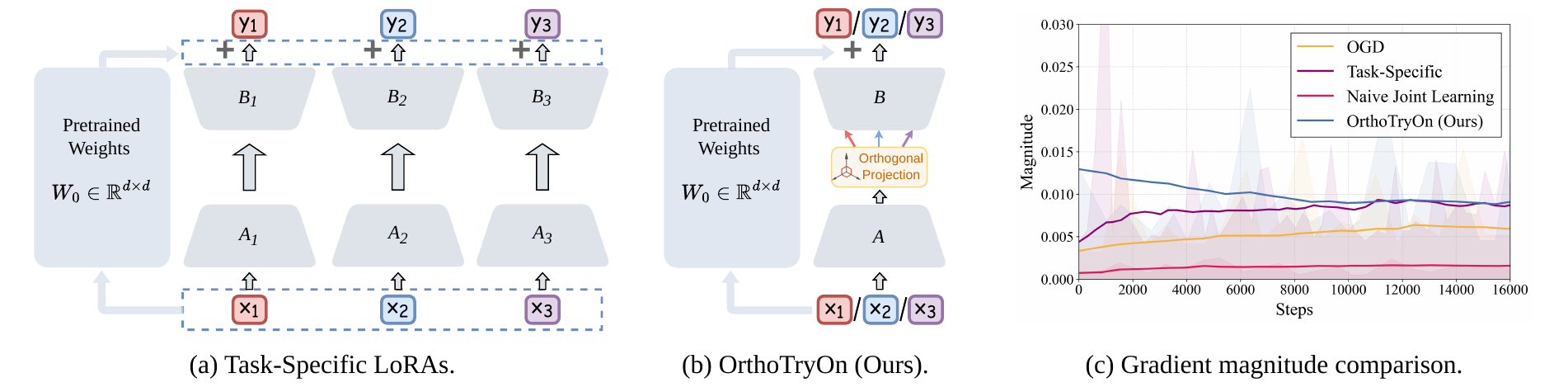}
        \put(88.7,21.7){\tiny \cite{ogd}}
    \end{overpic}
    \vspace{-4mm}
    \caption{(a) Task-Specific LoRAs maintain independent adapters for each task. (b) Our OrthoTryOn within a shared LoRA, enforcing decorrelated coordinate frames to mitigate inter-task conflict. (c) Gradient magnitudes of different methods across training steps, showing that OrthoTryOn effectively mitigates the gradient decay caused by inter-task conflict.}
	\vspace{-6mm}
    \label{fig:lora}
\end{figure}

Recent advances in diffusion-based image generation and editing have enabled increasingly controllable visual synthesis~\cite{dip,l2p,openvid,sourceswap,diffcod,ragd}, creating new opportunities for digital fashion. Among various fashion-related tasks, Virtual Try-On (VTON) aims to synthesize realistic images of a person wearing a target garment. Despite achieving impressive visual fidelity, most existing methods~\cite{idm-vton,d4-vton,hr-vton} remain specialized solutions constrained by stringent inputs (\eg, paired data or clean garment templates). Furthermore, current frameworks are typically designed to address a single predefined task. Consequently, supporting multiple fashion applications requires maintaining an ensemble of independently trained, task-specific Low-Rank Adaptation (LoRA) modules (as illustrated in Fig.~\ref{fig:lora}(a)), which inherently lacks scalability and poses significant deployment burdens.

To break the limitations of single-task specificity, pioneering works such as Any2AnyTryon~\cite{any2anytryon} and UniFit~\cite{unifit} have attempted to construct a unified generation paradigm. In such frameworks, adopting a single shared LoRA module for multi-task joint learning is a natural choice to maintain computational efficiency. However, this naive parameter sharing inevitably leads to noticeable performance degradation. Because different tasks possess distinct semantic objectives (\eg, spatial alignment for VTON \vs structural preservation for reconstruction), forcing them into an identical parameter space makes it difficult for the model to capture task-specific subtle differences.

In this work, we point out that the fundamental cause of performance degradation in unified fashion generation lies in the \textit{gradient conflict} during the multi-task joint optimization process. Through empirical analysis of gradient magnitudes within the shared LoRA parameters, we observe a sharp decay under naive multi-task training, as depicted in Fig.~\ref{fig:lora}(c), suggesting destructive interference among task gradients in the low-rank parameter space. Such interference drives the model toward a compromised solution that is suboptimal for all tasks.

To address these challenges, we propose \textbf{OrthoTryOn}, a novel unified framework that structurally mitigates negative transfer within a single shared LoRA module. Specifically, we design the Orthogonal Subspace Projection (OSP) strategy (Fig.~\ref{fig:lora}(b)), which introduces \emph{task-specific orthogonal rotations} $Q_i$ into the shared LoRA bottleneck. By rotating task-specific bottleneck features into decorrelated coordinate frames (without changing their magnitudes), OSP eliminates expected correlations between weight increments of different tasks and substantially reduces gradient-level interference, enabling stable joint training in a shared low-rank parameter space. In practice, each $Q_i$ is sampled once and then frozen, introducing negligible overhead.

Despite the statistical decorrelation introduced by OSP, residual semantic coupling may persist when the LoRA bottleneck dimension is small. To further enhance task discrimination at inference time, we introduce Fisher-guided Negative Guidance (FNG). FNG is a parameter-free strategy that quantifies inter-task sensitivity overlap using Fisher information and identifies the most confusable task as a hard negative condition within the Classifier-Free Guidance (CFG) framework~\cite{cfg}. OSP and FNG are highly synergistic: the former reduces gradient conflict during training, while the latter explicitly mitigates residual semantic leakage during generation.

Comprehensive experiments validate OrthoTryOn’s effectiveness for multi-task fashion generation. Unlike conventional unified frameworks, OrthoTryOn not only avoids the performance degradation caused by negative transfer, but also consistently outperforms independently trained task-specific models by preserving cross-task knowledge sharing. Furthermore, it demonstrates strong cross-architecture generalizability as a universal plug-in. The key contributions of this work are summarized as follows:
\begin{itemize}
\item We propose OrthoTryOn with Orthogonal Subspace Projection (OSP), leveraging task-specific orthogonal rotations $Q_i$ to construct decorrelated low-rank coordinate frames, enabling decorrelated joint optimization in expectation.
\item We introduce Fisher-guided Negative Guidance (FNG), a parameter-free inference strategy that utilizes diagonal Fisher sensitivity to quantify inter-task overlap and handle residual coupling.
\item OrthoTryOn achieves new state-of-the-art results across multiple benchmarks and consistently outperforms independently trained task-specific models, highlighting that properly structured parameter geometry can simultaneously suppress negative transfer and promote positive transfer in multi-task generation.
\end{itemize}
\section{Related Work}

\noindent\textbf{Fashion Image Generation.} Alongside recent progress in controllable image generation and editing~\cite{oneforall,personamagic,dvar,dvdpec,dciico,latexblend}, Virtual Try-On (VTON) has garnered significant attention due to its immense commercial potential~\cite{viton,hr-vton}. VITON~\cite{viton} pioneered image-based synthesis in this domain, while subsequent flow-based approaches~\cite{vitonhd,gp-vton,d4-vton} improved clothing-body alignment via dense appearance flow estimation. Recently, OmniVTON~\cite{omnivton} extended this paradigm to unconstrained real-world scenarios. However, mainstream VTON often relies on strict inputs, such as clean flat-lay garments. To bypass this, garment reconstruction~\cite{tryoffdiff,tryoffanyone} aims to extract standardized garments directly from human images. Concurrently, pose transfer~\cite{nted,cocosnet,sharpose} synthesizes novel poses while preserving identity and appearance, which is closely related to appearance-consistent correspondence under large deformations studied in visual tracking~\cite{hu2025exploiting,ding2026adaptive}.

To bridge these tasks, Any2AnyTryon~\cite{any2anytryon} and UniFit~\cite{unifit} adopt a shared parameter space for unified fashion generation. However, forcing a single shared parameter space to jointly fit tasks with substantial semantic discrepancies inevitably induces severe inter-task gradient conflicts. OrthoTryOn mitigates this issue through task-specific orthogonal rotations in the shared LoRA bottleneck, enabling more effective utilization of large-scale multi-task data.

\noindent\textbf{Task Decoupling in Deep Learning.} 
Orthogonality has long been used to stabilize optimization via weight initialization or manifold constraints~\cite{arjovsky2016unitary,lezcano2019cheap,wisdom2016full,saxe2013exact,mishkin2015all}, while contrastive learning improves discriminability by separating semantically confusable representations~\cite{c2p,deshadowmamba,bi2025dual}. More recently, projection-based methods such as PCGrad~\cite{pcgrad} and OGD~\cite{ogd} resolve conflicts by dynamically projecting gradients during backpropagation. However, these post-hoc manipulations require task-wise gradient isolation and additional backward passes, introducing non-negligible computational overhead. Moreover, explicitly discarding conflicting components may overly constrain optimization directions and suppress effective gradient magnitude.

In unified fashion generation, multiple tasks~\cite{any2anytryon,unifit} are typically accommodated within a single shared parameter space and distinguished only via conditional inputs. While computationally efficient, such tightly coupled parameter sharing inevitably induces severe inter-task gradient interference, leading to suboptimal convergence. In contrast, OrthoTryOn adopts a forward architectural orthogonalization strategy within a shared low-rank space by inserting task-specific orthogonal rotations, structurally decorrelating task updates in expectation and avoiding costly gradient manipulation, while Fisher-guided Negative Guidance further mitigates residual semantic leakage at inference.

\noindent\phantomsection\label{rela_sec:lora}\textbf{Parameter-Efficient Fine-Tuning.} As foundation models scale up rapidly, full-parameter fine-tuning becomes computationally prohibitive and prone to overfitting on downstream tasks. To alleviate this, Parameter-Efficient Fine-Tuning (PEFT) methods~\cite{adapter,prefixtuning,prompttuning} have emerged, aiming to achieve comparable performance to full fine-tuning by updating only a minuscule fraction of parameters.

Among various PEFT techniques, Low-Rank Adaptation (LoRA)~\cite{lora} is widely adopted due to its exceptional effectiveness and zero additional inference latency. For a pre-trained linear weight $W_0 \in \mathbb{R}^{d_{in} \times d_{out}}$ and an input feature $x$, LoRA freezes $W_0$ and introduces a trainable low-rank bypass. The forward propagation is formulated as:
\begin{equation}
    y = x W_0 + \alpha x A B,
\end{equation}
where $A \in \mathbb{R}^{d_{in} \times r}$ and $B \in \mathbb{R}^{r \times d_{out}}$ are the down- and up-projection matrices, respectively, with a bottleneck rank $r \ll \min(d_{in}, d_{out})$. The hyperparameter $\alpha$ scales the low-rank module and is omitted in subsequent derivations for simplicity.

Typically, $A$ is initialized with a Gaussian distribution and $B$ with zeros, ensuring the initial bypass output is zero to perfectly preserve pre-trained representations. While extensively injecting LoRA modules into Transformer layers (\eg, attention and feed-forward networks) maximizes fitting capacity for multi-task generation, directly employing a shared LoRA space for joint learning inevitably triggers severe inter-task gradient conflicts, as revealed in Sec.~\ref{method-sec:motivation}.
\section{Methods}\label{sec:Methods}

\begin{figure}[t]
	\centering
	\includegraphics[width=\textwidth]{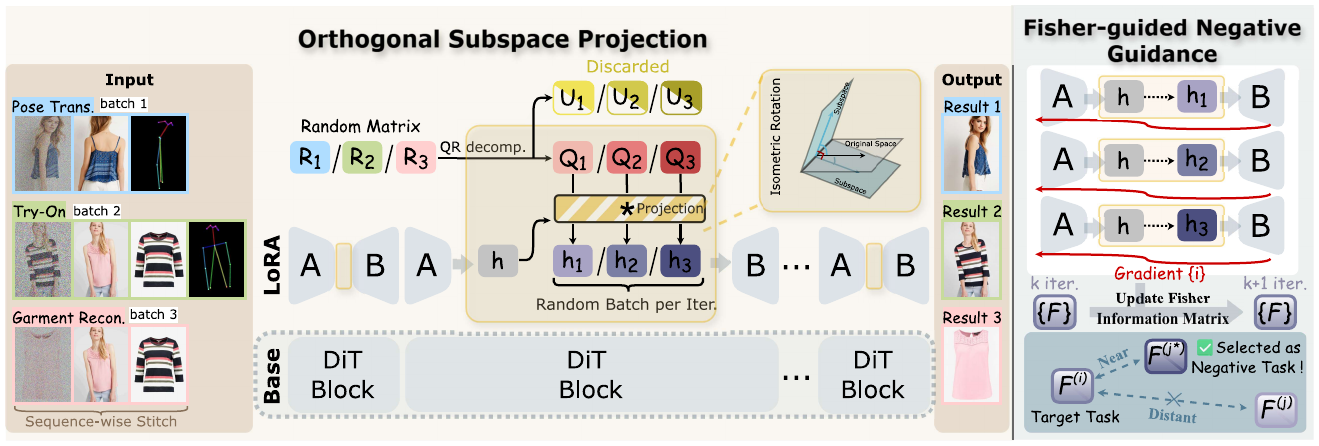}
    \vspace{-2mm}
    \caption{Overview of OrthoTryOn. Orthogonal Subspace Projection utilizes task-specific orthogonal matrices $Q_i$ in the shared LoRA module to rotate bottleneck features into decorrelated coordinate frames, achieving statistically decorrelated optimization in expectation. Based on parameter sensitivity tracked during training, Fisher-guided Negative Guidance explicitly suppresses the interfering task via CFG to prevent semantic leakage.}
	\vspace{-4mm}
    \label{fig:pipeline}
\end{figure}

\subsection{Overview of Universal Fashion Generation}\label{method-sec:overview}
We present a universal generative paradigm for multi-task fashion image editing. Rather than devising task-specific subnetworks, we seamlessly concatenate visual conditions along the sequence dimension: virtual try-on requires a reference model, a target garment, and a pose skeleton map; garment reconstruction utilizes a reference model and a reference garment (providing background attributes); and pose transfer relies on a reference model and a target pose skeleton map. Guided by task-specific text prompts, this concatenated sequence is directly fed into a unified Diffusion Transformer backbone.

While this paradigm unifies tasks at the input level, employing standard LoRA for joint multi-task learning inevitably encounters a parameter optimization bottleneck due to gradient conflicts (elaborated in Sec.~\ref{method-sec:motivation}). To overcome this, we propose the OrthoTryOn framework, comprising Orthogonal Subspace Projection (OSP) and Fisher-guided Negative Guidance (FNG), which enables architecture-agnostic multi-task decoupling and synergistic optimization within the shared LoRA space.

\subsection{Motivation}\label{method-sec:motivation}
To investigate the performance degradation in naive joint multi-task learning, we analyze the gradient dynamics during optimization. Our empirical analysis suggests that inter-task gradient interference is a major source of the model's performance deterioration. Specifically, we track and quantify the $L_2$ norm of the backpropagated gradients for the virtual try-on task under two settings: single-task training and joint multi-task learning. As illustrated in Fig.~\ref{fig:lora}(c), the gradient norm under single-task fine-tuning stabilizes at approximately $8.5 \times 10^{-3}$, whereas under joint multi-task optimization, it is reduced to about one-fifth of the single-task baseline. This pronounced attenuation is consistent with destructive interactions among task updates in the shared parameter space. Because our training framework employs uniform task sampling, updates from semantically distinct tasks are alternated throughout training and may partially counteract one another. Consequently, the shared parameters are driven toward a compromise that can be suboptimal for individual tasks.

\subsection{Orthogonal Subspace Projection}\label{method-sec:osp}
To resolve the gradient cancellation bottleneck, we propose Orthogonal Subspace Projection (OSP), which rotates task-specific features into decorrelated coordinate frames within the shared low-rank bottleneck to minimize inter-task interference.

Formally, suppose we jointly optimize $N$ fashion generation tasks within a shared LoRA parameter space $\theta$, with $\mathcal{L}_i$ denoting the loss of the $i$-th task. An ideal multi-task optimization would minimize the joint objective while keeping cross-task gradients orthogonal:
\begin{equation}
\min_{\theta}\sum_{i=1}^{N}\mathcal{L}_{i}(\theta) \quad \text{s.t.} \quad \langle\nabla_{\theta}\mathcal{L}_{i},\nabla_{\theta}\mathcal{L}_{j}\rangle=0, \forall i\neq j,
\end{equation}
where $\langle\cdot,\cdot\rangle$ denotes the inner product. Rather than explicitly enforcing this constraint, which would require costly per-task gradient isolation and projection, we pursue a forward reparameterization that reduces expected cross-task gradient correlation through architectural design.

\noindent\textbf{Task-specific orthogonal rotations in LoRA.} Standard LoRA computes the weight increment $\Delta W = AB$ via a low-rank down-projection $A \in \mathbb{R}^{d_{in} \times r}$ and up-projection $B \in \mathbb{R}^{r \times d_{out}}$. In OSP, for each task $i \in \{1, \dots, N\}$, we interpose a task-specific orthogonal matrix $Q_i \in \mathbb{R}^{r \times r}$ between $A$ and $B$, where $Q_i^\top Q_i = I$. The forward pass for an input feature $x$ becomes:
\begin{equation}\label{eq:osp_forward}
    y = x W_0 + x A Q_i B.
\end{equation}
Intuitively, $Q_i$ performs an \emph{isometric rotation} in the bottleneck subspace, assigning each task a distinct coordinate frame without changing feature magnitudes.

\noindent\textbf{Weight increment decorrelation.} The weight increment of task $i$ is $\Delta W_i = A Q_i B$. For distinct tasks $i \neq j$, their Frobenius inner product is:
\begin{equation}\label{eq:weight_decorrelation}
\langle \Delta W_i, \Delta W_j \rangle_F = \operatorname{tr}(B^\top Q_i^\top A^\top A Q_j B).
\end{equation}
When $Q_j$ is sampled independently from the Haar measure on $\mathcal{O}(r)$, symmetry implies $\mathbb{E}[Q_j]=\mathbf{0}$ (for $r\ge 2$). Therefore,
\begin{equation}
\mathbb{E}_{Q_j} \!\left[ \langle \Delta W_i, \Delta W_j \rangle_F \right]
= \operatorname{tr}\!\left( B^\top Q_i^\top A^\top A \cdot \mathbb{E}[Q_j] \cdot B \right) = 0.
\end{equation}
Thus, OSP achieves \emph{exact} decorrelation of cross-task weight increments in expectation.

\noindent\textbf{Gradient interference analysis.} Let $G_i = \frac{\partial \mathcal{L}_i}{\partial (A Q_i B)}$ denote the gradient of the loss with respect to the weight increment. By the chain rule,
\begin{equation}
\nabla_{B} \mathcal{L}_i = Q_i^\top A^\top G_i,
\qquad
\nabla_{A} \mathcal{L}_i = G_i B^\top Q_i^\top.
\end{equation}
We analyze cross-task interference under a \textit{single-step} setting: at any given iteration, $A$ and $B$ are fixed from the previous update, so $G_i$ depends on $Q_i$ (through Eq.~\ref{eq:osp_forward}) but is \textit{functionally independent of} $Q_j$ for $j \neq i$. Taking the cross-task gradient inner product on $B$ as a representative case:
\begin{equation}\label{eq:grad_inner_B}
\langle \nabla_B \mathcal{L}_i, \nabla_B \mathcal{L}_j \rangle
= \operatorname{tr}(G_i^\top A Q_i Q_j^\top A^\top G_j),
\end{equation}
and analogously $\langle \nabla_A \mathcal{L}_i, \nabla_A \mathcal{L}_j \rangle = \operatorname{tr}(Q_i B G_i^\top G_j B^\top Q_j^\top)$ for parameter $A$. We state the unified result below.

\begin{property}\label{prop:gradient_suppression}
Assume the loss function $\mathcal{L}$ is twice differentiable. At any optimization step conditional on fixed $A$ and $B$, let the task-specific rotations $Q_i, Q_j \in \mathcal{O}(r)$ be independently sampled from the Haar measure. Then the expected cross-task gradient inner product satisfies, for both parameters $A$ and $B$:
\begin{equation}\label{eq:grad_suppression}
\left| \mathbb{E}_{Q_i, Q_j}\!\left[\langle \nabla \mathcal{L}_i, \nabla \mathcal{L}_j \rangle \mid A, B \right] \right|
\le \mathcal{O}(1/r) \cdot C(A, B, G, \mathcal{H}),
\end{equation}
\end{property}
where $C$ is a constant determined by the parameter norms, per-step gradient scales, and bounded Hessian approximations, independent of the rank $r$.

Unlike the weight-increment case, this gradient bound relaxes to $\mathcal{O}(1/r)$ because $G_j$ functionally depends on $Q_j$ via the forward pass. While Haar symmetry ($\mathbb{E}[Q_j]=\mathbf{0}$) eliminates the first-order interference, the remaining second-order coupling stems from the Hessian operator. Applying the Haar-orthogonal conjugation identity to this term reveals the exact $1/r$ decay rate. Detailed proofs are provided in the supplementary material.

\noindent\textbf{Sampling and freezing orthogonal rotations.}
Each $Q_i$ is generated once per LoRA layer and per task, and remains frozen throughout training. In practice, we sample a Gaussian matrix and apply QR decomposition, retaining the orthogonal factor as $Q_i$. Since $Q_i^\top Q_i = I$, OSP is isometric by construction ($\|hQ_i\|_2=\|h\|_2$), avoiding spectral scaling and anisotropic distortion that may arise from unnormalized random projections. The storage footprint is negligible: only a fixed seed is required.

\subsection{Fisher-guided Negative Guidance}\label{method-sec:fng}
While OSP reduces expected cross-task gradient correlation, the $\mathcal{O}(1/r)$ suppression factor indicates that non-negligible residual coupling may persist, particularly within highly compressed low-rank bottlenecks (\eg, $r=4$). When tasks share similar visual semantics, this residual overlap in parameter sensitivities can induce coupled conditional velocity field predictions during inference. Intuitively, relying on these overlapping parameter subsets causes the learned vector fields to exhibit correlated directional biases, increasing the risk of semantic entanglement at generation time. To handle this residual semantic leakage, we introduce Fisher-guided Negative Guidance (FNG), a plug-and-play inference strategy designed to operate synergistically with OSP.

The core philosophy of FNG is to proactively identify the most severe ``interfering task'' and suppress it during the decoding phase. Instead of relying on feature-space similarity, we approximate inter-task sensitivity overlap using parameter-space statistics accumulated during training. Specifically, following standard practices~\cite{ewc,online-ewc,prlf}, we implement an empirical-Fisher-style proxy for per-parameter sensitivity. Since computing the full Fisher Information Matrix is intractable for large-scale models, we use its diagonal approximation to characterize the sensitivity of the $i$-th task to specific parameters.

To capture stable task sensitivities and mitigate early-stage gradient noise, we maintain an exponential moving average of squared gradients for each task to estimate its diagonal empirical-Fisher-style proxy $F^{(i)}$. At any given iteration step $k$, the vector is efficiently updated as:
\begin{equation}
F^{(i)}|_{k} = \beta \cdot F^{(i)}|_{k-1} + (1 - \beta) \cdot \mathbb{E}\left[ (\nabla_{\theta} \mathcal{L}_i)^2 \right],
\end{equation}
where $\beta \in [0, 1)$ is the momentum coefficient, the expectation is taken over the current mini-batch, $\theta$ denotes the trainable LoRA parameters across all adapted layers, and $F^{(i)}$ is updated only when task $i$ is sampled. This tracked statistic reflects how strongly each parameter contributes to optimizing a specific task. After training, fully converged sensitivity vectors are used only offline to identify each task’s most interfering task $j^*$. The high-dimensional Fisher vectors are then discarded, leaving only a discrete $i \mapsto j^*$ mapping for inference, which requires negligible storage and no trainable parameters.

During inference for task $i$, we identify the task with the highest Fisher similarity $j^*$ via cosine similarity between Fisher vectors, computed as $j^* = \arg\max_{j \neq i} S(i, j)$, where:
\begin{equation}
S(i, j) = \frac{\langle F^{(i)}, F^{(j)} \rangle}{\|F^{(i)}\|_2 \|F^{(j)}\|_2}.
\end{equation}

While multi-negative guidance is theoretically possible, adopting the single most severe interferer provides an optimal balance between disambiguation efficacy and computational cost. We then modify the conditional velocity prediction by replacing the unconditional null-prompt in standard Classifier-Free Guidance with the explicit condition of the interfering task:
\begin{equation} 
\hat{v} = v(x_t, t, c_{j^*}) + s \cdot \Big( v(x_t, t, c_i) - v(x_t, t, c_{j^*}) \Big),
\end{equation}
where $c_i$ and $c_{j^*}$ represent the condition inputs for the target and the hard negative tasks, respectively, and $s$ is the guidance scale. In the learned conditional velocity field, this operation explicitly pushes the generation trajectory away from the most heavily coupled semantic sub-manifold. Geometrically, this modifies the local vector field by introducing a repulsive component along the most correlated task direction, while preserving attraction toward the desired conditional manifold. Benefiting from this design, FNG effectively reduces semantic leakage without introducing any trainable parameters, mitigating task confusion in joint multi-task generation.
\section{Experiments}\label{sec:Experiments}
\begin{figure}[t]
	\centering
	\includegraphics[width=\textwidth]{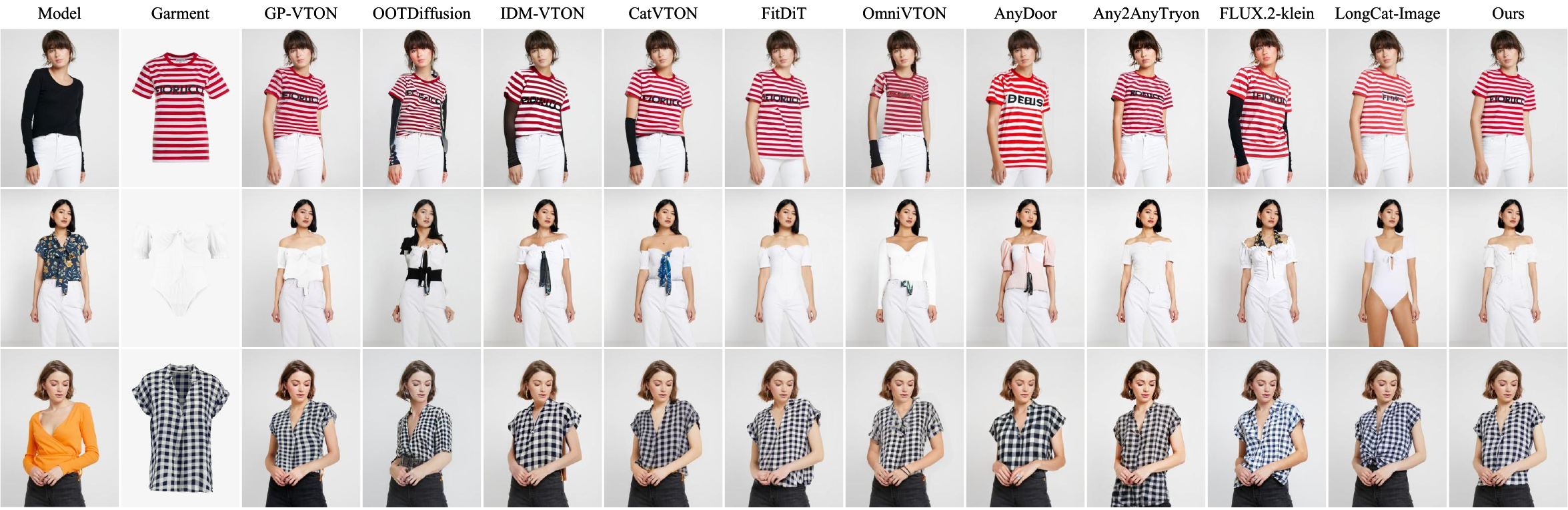}	
	\caption{Qualitative comparison of virtual try-on on VITON-HD dataset~\cite{vitonhd}.} 
	\label{fig:try-on}
    \vspace{-2mm}
\end{figure}

\subsection{Experimental Settings}
\noindent\textbf{Implementation Details.} We adopt LongCat-Image-Edit~\cite{longcat} as our backbone. All experiments are implemented in PyTorch 2.6.0 using four NVIDIA RTX A6000 GPUs. During joint training, we employ uniform task sampling and optimize the model for $10,000$ iterations using the AdamW optimizer~\cite{adamw} (batch size 16, base learning rate $1 \times 10^{-4}$ with $1,000$ warmup steps). The LoRA rank is fixed at 128. Input resolutions are set to $512 \times 384$ for virtual try-on and garment reconstruction, and $512 \times 352$ for pose transfer. For the dynamic diagonal Fisher Information Matrix estimation in FNG, the EMA momentum coefficient $\beta$ is set to 0.99. During inference, we use 50 sampling steps with an FNG scale of 2.0 for VTON and 1.5 for the remaining tasks.

\noindent\textbf{Datasets.} We jointly train our model on VITON-HD~\cite{vitonhd} and DeepFashion~\cite{deepfashion}. For virtual try-on and garment reconstruction, following Any2AnyTryon~\cite{any2anytryon}, we construct a shared subset from VITON-HD comprising 11,647 training and 2,032 testing quadruplets, from which the requisite condition-target tuples for each specific task are seamlessly extracted. For pose transfer, following the protocol in~\cite{progressive}, we partition DeepFashion into 101,966 training and 8,570 testing pairs depicting the same identity under different poses. Text conditions are generated via Qwen2.5-VL-7B-Instruct~\cite{qwen2}, and human skeletons are extracted using HRNet~\cite{hrnet}.

\begin{table}[!t]
  \small
  \centering
  \setlength{\abovecaptionskip}{0.2cm}
  \caption{Quantitative comparison across virtual try-on, garment reconstruction, and pose transfer. Each task is evaluated using four core metrics. The best and second-best results are highlighted in \textbf{bold} and \underline{underline}, respectively. Missing values for single-task experts are denoted with \textcolor{gray}{-}.}
  \label{tab:main_results}
  \resizebox{\textwidth}{!}{
  \begin{tabular}{l | c | cccc | cccc | cccc}
    \toprule
    \multirow{2}{*}{\textbf{Method}} & \multirow{2}{*}{\textbf{Capability}} & \multicolumn{4}{c|}{\textbf{Virtual Try-On}} & \multicolumn{4}{c|}{\textbf{Garment Recon.}} & \multicolumn{4}{c}{\textbf{Pose Transfer}} \\
    \cmidrule(lr){3-6} \cmidrule(lr){7-10} \cmidrule(lr){11-14}
    & & \textbf{LPIPS}↓ & \textbf{SSIM}↑ & \textbf{FID}↓ & \textbf{KID}↓ & \textbf{LPIPS}↓ & \textbf{DISTS}↓ & \textbf{FID}↓ & \textbf{CLIP-I}↑ & \textbf{LPIPS}↓ & \textbf{SSIM}↑ & \textbf{FID}↓ & \textbf{CLIP-I}↑ \\
    \midrule
    GP-VTON~\cite{gp-vton} & VTON Only & 0.068 & \underline{0.872} & 11.708 & 3.990 & \textcolor{gray}{-} & \textcolor{gray}{-} & \textcolor{gray}{-} & \textcolor{gray}{-} & \textcolor{gray}{-} & \textcolor{gray}{-} & \textcolor{gray}{-} & \textcolor{gray}{-} \\
    OOTDiffusion~\cite{ootd} & VTON Only & 0.132 & 0.784 & 15.136 & 5.774 & \textcolor{gray}{-} & \textcolor{gray}{-} & \textcolor{gray}{-} & \textcolor{gray}{-} & \textcolor{gray}{-} & \textcolor{gray}{-} & \textcolor{gray}{-} & \textcolor{gray}{-} \\
    IDM-VTON~\cite{idm-vton} & VTON Only & 0.082 & 0.816 & 10.745 & 2.229 & \textcolor{gray}{-} & \textcolor{gray}{-} & \textcolor{gray}{-} & \textcolor{gray}{-} & \textcolor{gray}{-} & \textcolor{gray}{-} & \textcolor{gray}{-} & \textcolor{gray}{-} \\
    CatVTON~\cite{catvton} & VTON Only & \textbf{0.057} & 0.870 & \underline{9.015} & \underline{1.091} & \textcolor{gray}{-} & \textcolor{gray}{-} & \textcolor{gray}{-} & \textcolor{gray}{-} & \textcolor{gray}{-} & \textcolor{gray}{-} & \textcolor{gray}{-} & \textcolor{gray}{-} \\
    FitDiT~\cite{fitdit} & VTON Only & 0.106 & 0.830 & 10.340 & 1.648 & \textcolor{gray}{-} & \textcolor{gray}{-} & \textcolor{gray}{-} & \textcolor{gray}{-} & \textcolor{gray}{-} & \textcolor{gray}{-} & \textcolor{gray}{-} & \textcolor{gray}{-} \\
    OmniVTON~\cite{omnivton} & VTON Only & 0.145 & 0.832 & 9.621 & 1.323 & \textcolor{gray}{-} & \textcolor{gray}{-} & \textcolor{gray}{-} & \textcolor{gray}{-} & \textcolor{gray}{-} & \textcolor{gray}{-} & \textcolor{gray}{-} & \textcolor{gray}{-} \\
    
    TryOffDiff~\cite{tryoffdiff} & Recon. Only & \textcolor{gray}{-} & \textcolor{gray}{-} & \textcolor{gray}{-} & \textcolor{gray}{-} & \underline{0.212} & 0.227 & 15.273 & 0.884 & \textcolor{gray}{-} & \textcolor{gray}{-} & \textcolor{gray}{-} & \textcolor{gray}{-} \\
    TryOffAnyone~\cite{tryoffanyone} & Recon. Only & \textcolor{gray}{-} & \textcolor{gray}{-} & \textcolor{gray}{-} & \textcolor{gray}{-} & 0.245 & \underline{0.211} & 11.488 & 0.904 & \textcolor{gray}{-} & \textcolor{gray}{-} & \textcolor{gray}{-} & \textcolor{gray}{-} \\
    
    CoCosNet-v2~\cite{cocosnet} & Pose Only & \textcolor{gray}{-} & \textcolor{gray}{-} & \textcolor{gray}{-} & \textcolor{gray}{-} & \textcolor{gray}{-} & \textcolor{gray}{-} & \textcolor{gray}{-} & \textcolor{gray}{-} & 0.227 & 0.724 & 13.325 & \textcolor{gray}{-} \\
    NTED~\cite{nted} & Pose Only & \textcolor{gray}{-} & \textcolor{gray}{-} & \textcolor{gray}{-} & \textcolor{gray}{-} & \textcolor{gray}{-} & \textcolor{gray}{-} & \textcolor{gray}{-} & \textcolor{gray}{-} & 0.200 & 0.736 & 7.645 & 0.916 \\
    PoCoLD~\cite{pocold} & Pose Only & \textcolor{gray}{-} & \textcolor{gray}{-} & \textcolor{gray}{-} & \textcolor{gray}{-} & \textcolor{gray}{-} & \textcolor{gray}{-} & \textcolor{gray}{-} & \textcolor{gray}{-} & 0.192 & 0.743 & 8.416 & \textcolor{gray}{-} \\
    CFLD~\cite{cfld} & Pose Only & \textcolor{gray}{-} & \textcolor{gray}{-} & \textcolor{gray}{-} & \textcolor{gray}{-} & \textcolor{gray}{-} & \textcolor{gray}{-} & \textcolor{gray}{-} & \textcolor{gray}{-} & 0.182 & \underline{0.748} & 7.149 & 0.922 \\
    MCLD~\cite{mcld} & Pose Only & \textcolor{gray}{-} & \textcolor{gray}{-} & \textcolor{gray}{-} & \textcolor{gray}{-} & \textcolor{gray}{-} & \textcolor{gray}{-} & \textcolor{gray}{-} & \textcolor{gray}{-} & 0.176 & \textbf{0.756} & \underline{7.079} & \underline{0.926} \\
    \midrule
    AnyDoor~\cite{anydoor} & Unified & 0.113 & 0.808 & 13.403 & 5.793 & 0.474 & 0.340 & 65.130 & 0.781 & 0.636 & 0.346 & 61.643 & 0.716 \\
    Any2AnyTryon~\cite{any2anytryon} & Unified & 0.077 & 0.846 & 10.143 & 2.668 & 0.250 & 0.218 & \underline{10.771} & \underline{0.907} & \underline{0.175} & 0.705 & 12.250 & 0.916 \\
    FLUX.2-klein~\cite{flux2} & Unified & 0.114 & 0.836 & 14.651 & 7.126 & 0.356 & 0.272 & 30.491 & 0.864 & 0.256 & 0.634 & 8.598 & 0.881 \\
    LongCat-Image-Edit~\cite{longcat} & Unified & 0.101 & 0.840 & 10.587 & 1.528 & 0.292 & 0.237 & 16.157 & 0.880 & 0.205 & 0.694 & 18.703 & 0.899 \\
    \rowcolor[HTML]{F2F2F2}
    OrthoTryOn (Ours) & Unified & \underline{0.064} & \textbf{0.876} & \textbf{8.312} & \textbf{0.532} & \textbf{0.192} & \textbf{0.191} & \textbf{9.563} & \textbf{0.931} & \textbf{0.146} & 0.728 & \textbf{6.364} & \textbf{0.936} \\
    \bottomrule
  \end{tabular}}
\end{table}

\begin{figure}[t]
	\centering
	\includegraphics[width=\textwidth]{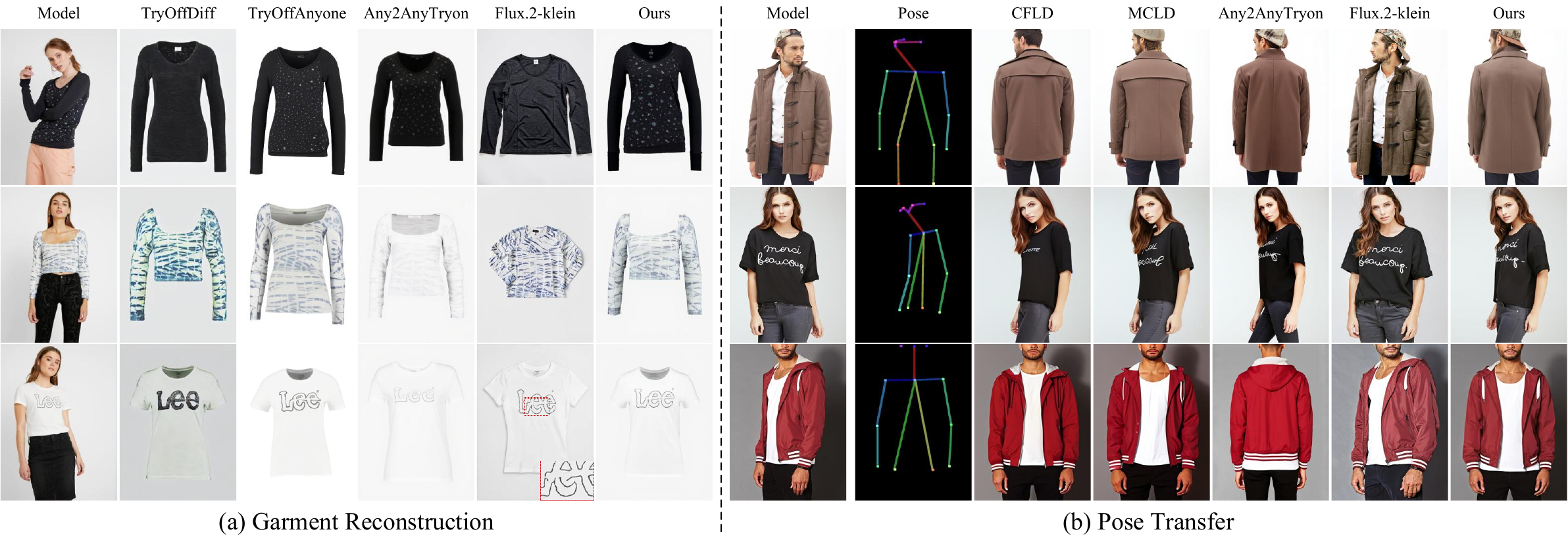}	
	\caption{Qualitative comparison of garment reconstruction on VITON-HD dataset~\cite{vitonhd} and pose transfer on DeepFashion dataset~\cite{deepfashion}. Please zoom in for better view.} 
	\label{fig:recon_pose}
    \vspace{-4mm}
\end{figure}

\subsection{Comparison with State-of-the-Art Methods}
To evaluate OrthoTryOn, we benchmark against four unified foundation models (AnyDoor~\cite{anydoor}, Any2AnyTryon~\cite{any2anytryon}, LongCat-Image-Edit, and FLUX.2-klein~\cite{flux2}) alongside numerous task-specific expert models. While the first three unified baselines are retrained on our multi-task dataset using their official implementations, we directly employ the pre-trained weights of FLUX.2-klein for zero-shot evaluation, taking full advantage of its inherent multi-conditional image generation capabilities without additional fine-tuning.

\noindent\textbf{Virtual Try-On.} We employ LPIPS~\cite{lpips} and SSIM~\cite{ssim} to evaluate perceptual quality and structural consistency, alongside FID~\cite{fid} and KID~\cite{kid} under an unpaired setting to simulate real-world scenarios. Beyond general baselines, we compare OrthoTryOn against six task-specific SOTA experts: GP-VTON~\cite{gp-vton}, OOTDiffusion~\cite{ootd}, IDM-VTON~\cite{idm-vton}, CatVTON~\cite{catvton}, FitDiT~\cite{fitdit}, and OmniVTON~\cite{omnivton}. Quantitative results in Tab.~\ref{tab:main_results} indicate that OrthoTryOn significantly outperforms all general baselines. Notably, despite being a unified framework, it even surpasses the majority of single-task experts. This superiority is largely attributed to OSP's effective disentanglement: by suppressing inter-task gradient conflicts at a rate that scales inversely with the bottleneck dimension, our model mitigates destructive interference during joint optimization, enabling stable and efficient utilization of large-scale multi-task data. Visual comparisons in Fig.~\ref{fig:try-on} demonstrate our method's exceptional garment fidelity, effectively preventing texture distortions and reducing residual artifacts from the original clothing. By operating under a shared parameter budget, OrthoTryOn highlights that structured parameter geometry can prevent the capacity dilution typically observed in naive multi-task learning.

\noindent\textbf{Garment Reconstruction.} We evaluate garment reconstruction using FID, LPIPS, CLIP-I~\cite{clip}, and DISTS~\cite{dists}. As detailed in Tab.~\ref{tab:main_results}, compared to general baselines and two task-specific experts (TryOffDiff~\cite{tryoffdiff}, TryOffAnyone~\cite{tryoffanyone}), OrthoTryOn achieves superior performance across all metrics. Notably, it excels in LPIPS and DISTS, indicating a highly accurate preservation of both global perceptual realism and local structural details. Despite the semantic gap between object-level garment reconstruction and human-centric generation, OSP successfully shields garment texture features from inter-task interference during joint optimization. As depicted in Fig.~\ref{fig:recon_pose} (a), unlike baselines that struggle with severe texture distortions and incorrect garment categorization, our method delivers accurate reconstruction, strictly preserving intricate textural patterns and complex topological structures.

\noindent\textbf{Pose Transfer.} For pose transfer, we employ FID, LPIPS, SSIM, and CLIP-I. We compare against five task-specific SOTAs: CoCosNet-v2~\cite{cocosnet} and PoCoLD~\cite{pocold} (reporting official results), alongside NTED~\cite{nted}, CFLD~\cite{cfld}, and MCLD~\cite{mcld} (evaluated using generated images released by the authors). According to the quantitative benchmarks, OrthoTryOn significantly outperforms all general baselines and surpasses experts across most metrics. Despite a slight SSIM trade-off (likely attributable to the use of dense spatial priors like DensePose~\cite{densepose} in some expert models, whereas our framework relies on sparse skeletal conditions), our method achieves a 0.715 FID improvement, indicating better alignment with real-world distributions. As shown in Fig.~\ref{fig:recon_pose} (b), OrthoTryOn infers high-fidelity target views from single-view inputs and strictly maintains fine-grained garment textures without artifacts or blurring, even under drastic pose variations.

\begin{table*}[!t]
  \small
  \centering
  \setlength{\abovecaptionskip}{0.2cm}
  \caption{Ablation study across three fashion generation tasks. Base (Joint-Learning): naive multi-task learning using a single shared LoRA. Base (Task-Specific): trained exclusively on individual tasks. OSP-R: inserting non-orthogonal random matrices in the LoRA bottleneck. FNG$^*$: guidance using the task with the lowest Fisher similarity.}
  \label{tab:ablation}
  \resizebox{\textwidth}{!}{ 
  \begin{tabular}{l | cccc | cccc | cccc}
    \toprule
    \multirow{2}{*}{\textbf{Setting}} & \multicolumn{4}{c|}{\textbf{Virtual Try-On}} & \multicolumn{4}{c|}{\textbf{Garment Recon.}} & \multicolumn{4}{c}{\textbf{Pose Transfer}} \\
    \cmidrule(lr){2-5} \cmidrule(lr){6-9} \cmidrule(lr){10-13}
    & \textbf{LPIPS}↓ & \textbf{SSIM}↑ & \textbf{FID}↓ & \textbf{KID}↓ & \textbf{LPIPS}↓ & \textbf{DISTS}↓ & \textbf{FID}↓ & \textbf{CLIP-I}↑ & \textbf{LPIPS}↓ & \textbf{SSIM}↑ & \textbf{FID}↓ & \textbf{CLIP-I}↑ \\
    \midrule
    (a) Base (Joint-Learning) & 0.101 & 0.840 & 10.587 & 1.528 & 0.292 & 0.237 & 16.157 & 0.880 & 0.205 & 0.694 & 18.703 & 0.899 \\
    (b) Base (Task-Specific) & 0.095 & 0.840 & 9.487 & 0.941 & 0.211 & \textbf{0.190} & 10.428 & 0.925 & 0.188 & 0.685 & 7.445 & 0.926 \\
    (c) Base + OSP-R & 0.114 & 0.812 & 9.318 & 1.014 & 0.491 & 0.347 & 20.412 & 0.790 & 0.388 & 0.602 & 15.423 & 0.873 \\
    (d) Base + OSP & 0.084 & 0.867 & 8.386 & 0.576 & 0.203 & 0.194 & 9.787 & 0.929 & 0.161 & 0.716 & 6.632 & 0.928 \\
    (e) Base + OSP + FNG$^*$ & \textbf{0.064} & \textbf{0.876} & 8.387 & 0.640 & 0.194 & 0.193 & 9.871 & 0.930 & 0.154 & 0.723 & 7.318 & 0.930 \\
    \midrule
    \rowcolor[HTML]{F2F2F2}
    \textbf{OrthoTryOn (Ours)} & \textbf{0.064} & \textbf{0.876} & \textbf{8.312} & \textbf{0.532} & \textbf{0.192} & 0.191 & \textbf{9.563} & \textbf{0.931} & \textbf{0.146} & \textbf{0.728} & \textbf{6.364} & \textbf{0.936} \\
    \bottomrule
  \end{tabular}
  }
\end{table*}

\begin{table*}[!t]
  \small
  \centering
  \setlength{\abovecaptionskip}{0.2cm}
  \caption{Quantitative evaluation of cross-architecture generalizability. For Any2AnyTryon, we only apply OSP without FNG, since its distilled FLUX.1-dev backbone already incorporates strong built-in CFG priors.}
  \label{tab:cross-backbone}
  
  \resizebox{\textwidth}{!}{ 
  \begin{tabular}{l | cccc | cccc | cccc}
    \toprule
    \multirow{2}{*}{\textbf{Setting}} & \multicolumn{4}{c|}{\textbf{Virtual Try-On}} & \multicolumn{4}{c|}{\textbf{Garment Recon.}} & \multicolumn{4}{c}{\textbf{Pose Transfer}} \\
    \cmidrule(lr){2-5} \cmidrule(lr){6-9} \cmidrule(lr){10-13}
    & \textbf{LPIPS}↓ & \textbf{SSIM}↑ & \textbf{FID}↓ & \textbf{KID}↓ & \textbf{LPIPS}↓ & \textbf{DISTS}↓ & \textbf{FID}↓ & \textbf{CLIP-I}↑ & \textbf{LPIPS}↓ & \textbf{SSIM}↑ & \textbf{FID}↓ & \textbf{CLIP-I}↑ \\
    \midrule
    Any2AnyTryon~\cite{any2anytryon} & 0.077 & 0.846 & 10.143 & 2.668 & 0.250 & 0.218 & 10.771 & 0.907 & 0.175 & 0.705 & 12.250 & 0.916 \\
    \rowcolor[HTML]{F2F2F2}
    Any2AnyTryon + OSP & \textbf{0.058} & \textbf{0.868} & \textbf{9.180} & \textbf{1.800} & \textbf{0.224} & \textbf{0.207} & \textbf{10.127} & \textbf{0.913} & \textbf{0.170} & \textbf{0.707} & \textbf{11.609} & \textbf{0.916} \\
    \midrule
    AnyDoor~\cite{anydoor} & \textbf{0.113} & 0.808 & 13.403 & 5.793 & 0.474 & 0.340 & 65.130 & 0.781 & 0.636 & 0.346 & 61.643 & 0.716 \\
    \rowcolor[HTML]{F2F2F2}
    AnyDoor + OSP & 0.114 & 0.808 & 13.249 & 5.768 & 0.373 & 0.278 & 21.604 & 0.852 & 0.593 & 0.359 & 52.489 & 0.772 \\
    \rowcolor[HTML]{F2F2F2}
    AnyDoor + OSP + FNG & \textbf{0.113} & \textbf{0.811} & \textbf{12.806} & \textbf{5.369} & \textbf{0.346} & \textbf{0.263} & \textbf{18.810} & \textbf{0.860} & \textbf{0.592} & \textbf{0.364} & \textbf{48.375} & \textbf{0.783} \\
    \bottomrule
  \end{tabular}
  }
\end{table*}

\begin{figure}[t]
    \centering
    \begin{minipage}[c]{0.62\textwidth}
        \centering
        \includegraphics[width=\linewidth]{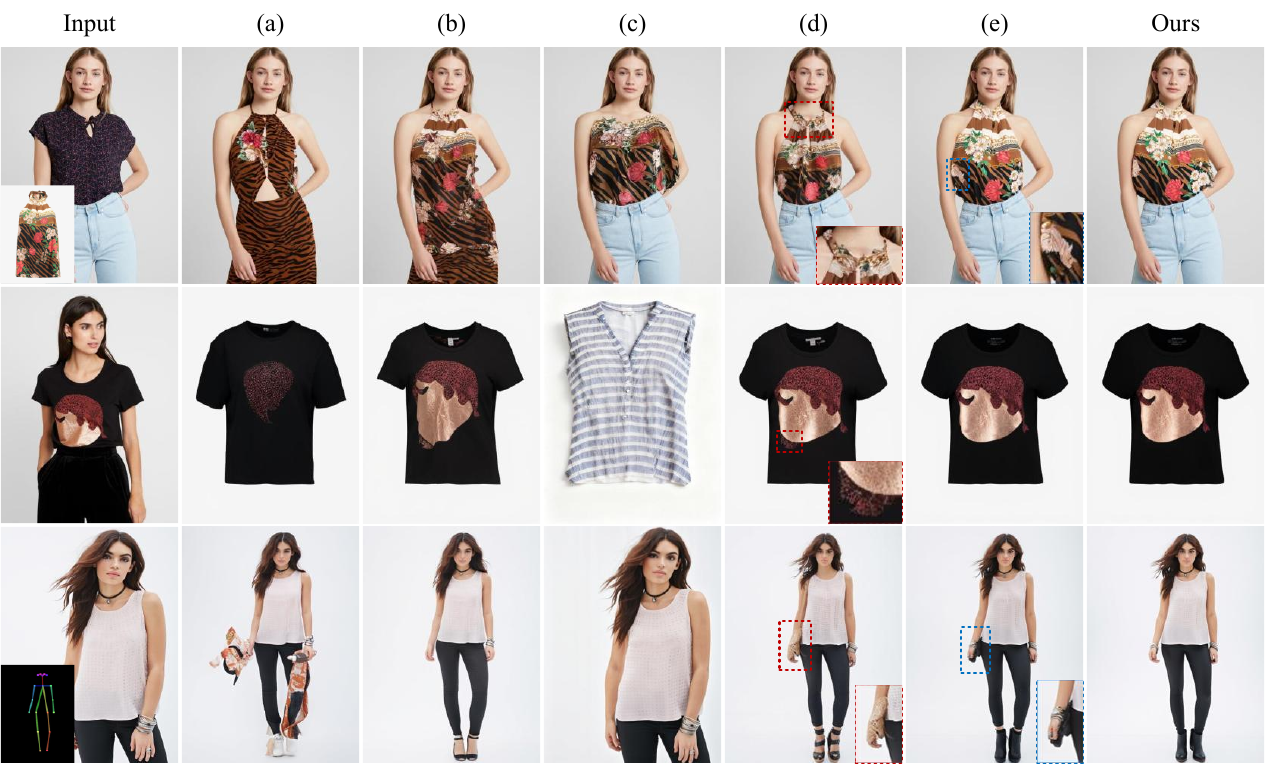}
        \vspace{-2mm}
        \caption{Qualitative ablation study on different variants. Rows from top to bottom: virtual try-on, garment reconstruction, and pose transfer.}
        \label{fig:ablation}
    \end{minipage}
    \hfill
    \begin{minipage}[c]{0.33\textwidth}
        \centering
        \includegraphics[width=\linewidth]{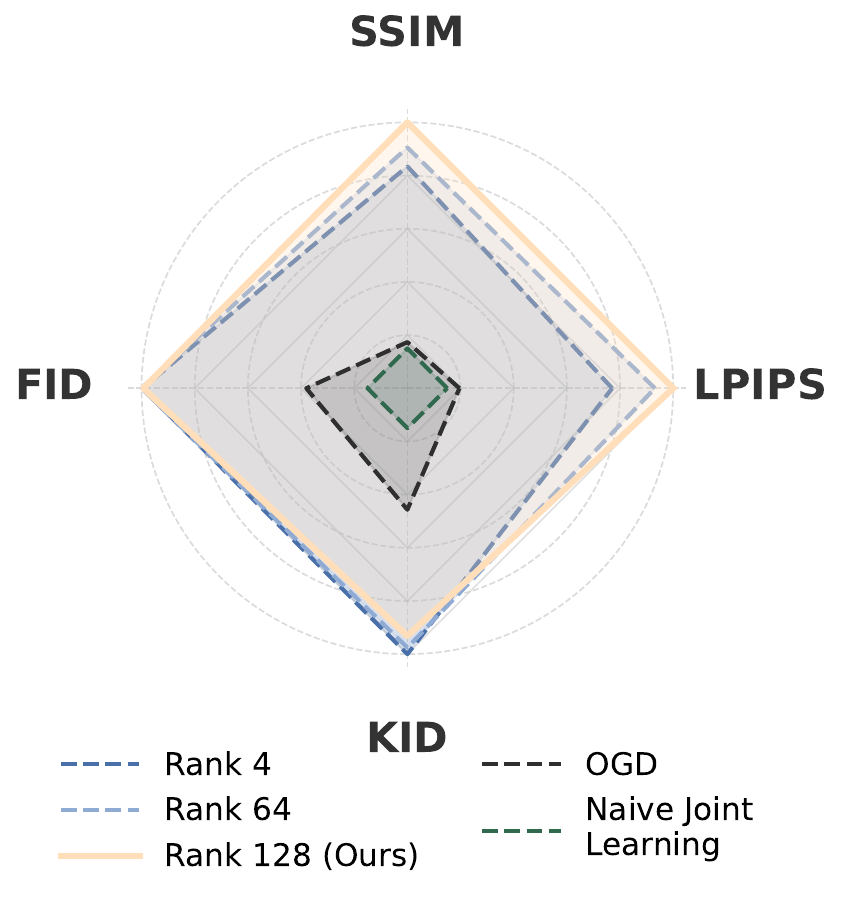}
        \vspace{-2mm}
        \caption{Performance comparison of different virtual try-on variants across multiple metrics.}
        \label{fig:radar}
    \end{minipage}
    \vspace{-4mm}
\end{figure}

\subsection{Ablation Study}
We conduct comprehensive ablations to validate our core components. Quantitative and qualitative results are summarized in Tab.~\ref{tab:ablation} and Fig.~\ref{fig:ablation}.

\noindent\textbf{Effectiveness of OSP.} Naive joint learning (variant (a)) suffers from severe artifacts and texture blur across all tasks, underperforming even the task-specific expert models (variant (b)). This degradation primarily stems from gradient conflicts and negative transfer among heterogeneous tasks. By introducing OSP (variant (d)), the model overcomes this bottleneck, outperforming both naive joint learning and expert models across most metrics. This superiority arises because OSP assigns each task a distinct low-rank coordinate frame while preserving feature norms, thereby reducing destructive cross-task interference yet retaining shared human and physical priors under stable joint optimization. To further verify that orthogonality is the key to stable decoupling, we replace $Q_i$ with non-orthogonal random matrices $R_i$ (variant (c)). This variant severely impairs generative capability: compared to variant (d), its FID scores on garment reconstruction and pose transfer deteriorate drastically by 10.625 and 8.791, respectively. Without the isometric constraint, random mixing introduces uncontrolled spectral scaling and anisotropic distortion in the bottleneck, which destabilizes optimization and amplifies negative transfer, ultimately leading to significant performance degradation.

\noindent\textbf{Effectiveness of FNG.} Although OSP effectively mitigates interference during joint learning, residual semantic coupling may persist under highly compressed low-rank bottlenecks, which can trigger semantic leakage during inference. As highlighted by the red boxes in Fig.~\ref{fig:ablation}, variant (d) relying solely on OSP still exhibits visual artifacts when generating local details. To address this, we introduce Fisher-guided Negative Guidance (FNG). We first evaluate a suboptimal variant FNG$^*$ (variant (e)), which forces the task with the lowest Fisher similarity (\ie, maximum semantic discrepancy) as the negative prompt. As shown in Tab.~\ref{tab:ablation}, while this strategy enhances structural alignment and eliminates the red-box artifacts, it introduces new visual flaws (blue boxes) and suffers from FID degradation. This likely occurs because the most distant task has minimal distributional overlap with the target task; thus, forced repulsion fails to accurately isolate genuine interfering features and instead injects biased perturbations. In contrast, OrthoTryOn selects the task with the highest Fisher similarity as the hard negative task. Utilizing the fully converged Fisher statistics tracked via EMA during training, this operation precisely targets highly confusable semantics. The final results indicate that the complete FNG module eradicates all visual flaws and achieves state-of-the-art performance across all metrics (\eg, boosting pose transfer FID to 6.364), demonstrating the necessity and effectiveness of repelling highly correlated tasks based on parameter sensitivity.

\noindent\textbf{Robustness of Orthogonal Decoupling.} To validate the robustness of OSP, we evaluate OrthoTryOn across varying LoRA ranks $r \in \{4, 64, 128\}$. As depicted in Fig.~\ref{fig:radar}, while reducing the rank to 4 yields a marginal decline in SSIM and LPIPS due to constrained parameter capacity, the overall generative fidelity (FID and KID) remains highly stable and significantly outperforms the naive joint learning baseline. This indicates that orthogonal coordinate frames remain beneficial even under extreme low-rank constraints. Furthermore, we benchmark against Orthogonal Gradient Descent (OGD), a representative post-hoc gradient projection method. Although OGD partially mitigates negative transfer compared to naive joint learning, discarding conflicting gradient components may hinder convergence by removing potentially useful optimization signals. Consequently, it is clearly worse than OrthoTryOn across all evaluation metrics.

\subsection{Cross-Architecture Generalizability}
To validate architectural generalizability, we adapt OrthoTryOn to Any2AnyTryon (FLUX.1-dev~\cite{flux}) and AnyDoor (Stable Diffusion 2.1~\cite{sd}). As shown in Tab.~\ref{tab:cross-backbone}, our components yield consistent gains across distinct generative paradigms.

For Any2AnyTryon, integrating OSP yields consistent improvements across all three evaluated tasks. In the Virtual Try-On setting, OSP significantly enhances fine-grained detail preservation and overall image realism, evidenced by a marked drop in LPIPS from 0.077 to 0.058 and a reduction in FID from 10.143 to 9.180. Similar gains are observed in Garment Reconstruction and Pose Transfer (\eg, Garment Recon. CLIP-I increases to 0.913 and Pose Transfer FID decreases to 11.609), indicating effective mitigation of inter-task gradient interference within the shared LoRA space. As noted, FNG is omitted in this setup due to its partial functional overlap with the distilled FLUX.1-dev backbone's built-in CFG priors. Nevertheless, OSP alone effectively stabilizes multi-task optimization by structurally reducing gradient conflicts in the shared low-rank space.

For AnyDoor, both OSP and FNG are seamlessly integrated. OrthoTryOn enables a substantial improvement in garment reconstruction, with FID dropping from 65.130 to 18.810. While the improvements in VTON are relatively modest, this is primarily because AnyDoor's pre-training already incorporates the VITON-HD dataset, leaving limited space for further optimization. Furthermore, although the absolute metrics for pose transfer remain suboptimal due to the inherent limitations of AnyDoor's local inpainting paradigm in handling large-scale spatial deformations, OrthoTryOn still achieves a substantial improvement in relative performance. This demonstrates that our framework effectively mitigates inter-task interference, allowing the backbone to better realize its multi-task capacity.

\section{Conclusion}
In this paper, we present OrthoTryOn, a highly effective framework for unified fashion generation that overcomes the negative transfer inherent in shared LoRA adaptation. By introducing Orthogonal Subspace Projection (OSP), we structurally enforce decorrelated coordinate frames via task-specific orthogonal rotations, which significantly suppresses gradient interference and enables stable joint optimization in expectation. Complementarily, Fisher-guided Negative Guidance (FNG) leverages empirical parameter sensitivities to explicitly mitigate residual semantic leakage during inference, without introducing any trainable parameters. Extensive experiments demonstrate that OrthoTryOn not only avoids the performance degradation typically observed in unified training, but also surpasses independently trained task-specific models, highlighting the importance of structured parameter geometry in unlocking effective multi-task generation. Moreover, OrthoTryOn generalizes robustly across diverse diffusion backbones, establishing itself as a universal plug-and-play adaptation mechanism.

\noindent\textbf{Limitation.} The $\mathcal{O}(1/r)$ interference bound of OSP implies that, under extremely small LoRA ranks coupled with many heterogeneous tasks, residual gradient coupling may become non-negligible. Although FNG partially compensates at inference time, it cannot fully recover training-stage information loss. In practice, moderately increasing the rank provides sufficient degrees of freedom to achieve better performance, as validated by our rank ablation study.

\section*{Acknowledgements}
This work was supported by the National Natural Science Foundation of China (NSFC) under Grant Nos. U24A20330, 62361166670, and 62406135; the Natural Science Foundation of Jiangsu Province under Grant No. BK20241198; and the Gusu Innovation and Entrepreneur Leading Talents Program under Grant No. ZXL2024362.
%
%
\bibliographystyle{splncs04}
\bibliography{main}
\end{document}